\pdfoutput=1

\documentclass[11pt]{article}

\usepackage[]{acl}

\usepackage{times}
\usepackage{latexsym}

\usepackage[T1]{fontenc}

\usepackage[utf8]{inputenc}

\usepackage{microtype}

\usepackage{booktabs}
\usepackage{times}
\usepackage{latexsym}
\usepackage{graphicx}
\usepackage{tabularx}
\usepackage{subcaption}
\title{Evaluating the Factuality of Zero-shot Summarizers \\ Across Varied Domains}

\author{Sanjana Ramprasad$^\diamondsuit$ \quad Kundan Krishna$^\clubsuit$ \quad  \textbf{Zachary C. Lipton}$^{\clubsuit}$ \quad \textbf{Byron C. Wallace}$^\diamondsuit$ 
  \\
$^\diamondsuit$Northeastern University \\
$^\clubsuit$ Carnegie Mellon University \\
\texttt{\small \{ramprasad.sa,b.wallace\}@northeastern.edu} \\
\texttt{\small \{kundank,zlipton\}@andrew.cmu.edu}}

\begin{document}
\maketitle
\begin{abstract}

Recent work has shown that large language models (LLMs) are capable of generating summaries \emph{zero-shot} (i.e., without explicit supervision) that 
are often comparable or even preferred to manually composed reference summaries. 
However, this prior work has focussed almost exclusively on evaluating news article summarization. How do zero-shot summarizers perform in other, potentially more specialized, domains?
In this work we evaluate zero-shot generated summaries across specialized domains including: biomedical articles, and legal bills (in addition to standard news benchmarks, for reference).
We focus especially on the factuality of outputs. 
We acquire annotations from domain experts to identify inconsistencies in summaries and systematically categorize these errors. 
We analyze whether the prevalence of a given domain in the pretraining corpus affects extractiveness and faithfulness of generated summaries of articles in this domain. 
We release all collected annotations to facilitate additional research toward measuring and realizing factually accurate summarization, beyond news articles.\footnote{The dataset can be downloaded from 
\url{https://github.com/sanjanaramprasad/zero_shot_faceval_domains}}

\end{abstract}

\section{Introduction}

Modern LLMs now offer strong zero-shot summarization performance, and even surpass fine-tuned models according to human assessments \cite{goyal2022news}.
Indeed, zero-shot summaries are sometimes deemed comparable in quality to reference summaries \cite{zhang2023benchmarking}.
Past evaluative work, however, has focused nearly exclusively on news article summarization, a domain in which there is no shortage of available training data.

But zero-shot summarization is perhaps \emph{most} appealing in niche domains where acquiring training data with which to fine-tune summarization models is sparse and may be prohibitively expensive to collect. 
Recent work \cite{shaib2023summarizing, tang2023evaluating} suggests the promise of zero-shot summarization in such domains.
However, there has not yet been a comprehensive investigation of the factuality of model outputs produced in zero-shot summarization across multiple domains (i.e., beyond news). 
Here we address this gap, and compare the quality of zero-shot summaries generated in niche domains (law, medicine) to those generated for news articles. 

In evaluating these models, we center the consistency and faithfulness of summaries generated by LLMs with respect to the input (source) document. 
Inconsistencies within summaries have long posed a challenge \citep{maynez2020faithfulness, pagnoni2021understanding}, 
motivating approaches 
intended to 
mitigate this issue \citep{zhu2020enhancing, cao2021cliff}, and 
for automated evaluation of factuality \citep{kryscinski2019evaluating, goyal2020evaluating, fabbri2021qafacteval, scialom2021questeval, laban2022summac, luo2023chatgpt}. 
Here we systematically assess the factual accuracy of zero-shot summarizers across a diverse set of specialized domains.

Specifically, we look to answer four major questions. 
(1) What is the \emph{prevalence} of errors in zero-shot summaries across various domains, and how does this compare to established results on news summarization tasks? 
(2) Are the \emph{types} of errors observed in these niche domains different from what has been seen in news article summarization?
(3) What is the relationship between the 
frequency of domains in training corpora and the likelihood of model hallucinations in these domains?
(4) Are existing automatic systems for factual evaluation reliable across multiple domains? 

To answer these questions, we enlist expert annotators to manually evaluate the outputs from two representative zero-shot summarization systems---GPT-3.5 ({\tt gpt-3.5-turbo-0301}; \citealt{brown2020language}) and Flan-T5-XL \citep{chung2022scaling}---across standard and niche summarization datasets.
Specifically, we evaluate (zero-shot) summaries of medical and legal documents, as well as news articles for reference. 

In general, we find that the proportion of factual inconsistencies in summaries varies considerably across domains, calling into question the community focus on news summarization datasets specifically. 
Further, we find evidence that the prevalence of articles in pretraining data from a given domain may correlate with the factuality of summaries of articles from the same. 
We speculate that this may be due to the model introducing content implicit in its weights in such cases (whereas it may have less ``knowledge'' in niche domains), although this would need to be validated in future work.

\section{Manual Evaluations of Summaries}

\paragraph{Data} 

We use XSUM \citep{narayan2018don} and CNN-DM \citep{hermann2015teaching} for news, as well as niche domains like PubMed (medicine; \citealt{cohan2018discourseaware}) and legal bills (law; \citealt{kornilova2019billsum}) for comparison. We select articles shorter than 4096 tokens from the test sets to accommodate model token limitations, resulting in approximately 22,000 articles for news, 3,000 for billsum, and 200 for PubMed. 
We randomly (i.i.d.) sample 50 articles from each domain.
We provide more data statistics in Appendix \ref{sec:appendix_summary_stats}
\paragraph{Model Details}
We run experiments with GPT-3.5 ({\tt gpt-3.5-turbo-0301}) and Flan-T5-XL \cite{chung2022scaling}. 
We use a general prompt similar to prior work \citep{goyal2022news} for generating summaries across domains. 
 Specifically, the prompt is as follows: "Article: [article]. Summarize the above article."
\paragraph{Annotation Collection}

\begin{figure*}
    \begin{subfigure}[b]{0.52\textwidth}
       
        \includegraphics[scale=0.5]{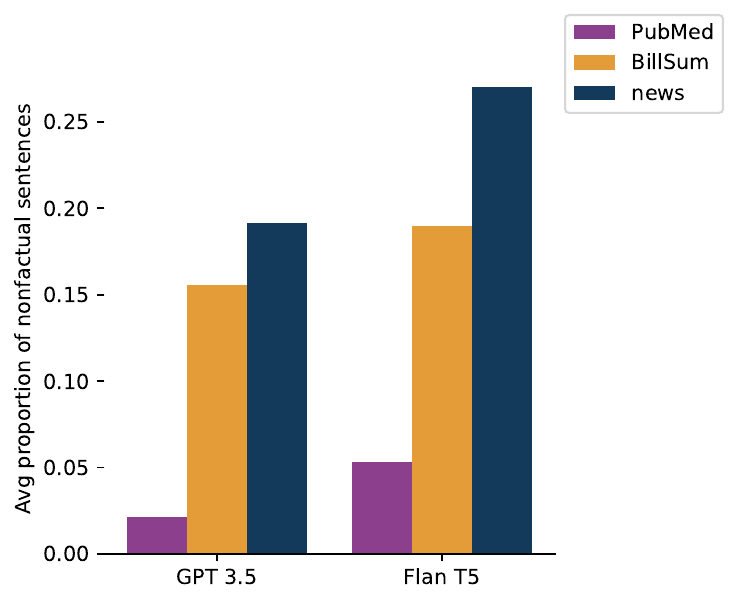}
        \caption{Prevalence of factual errors in each of domains}
        \label{fig:figure1}
    \end{subfigure}
    \hfill
    \begin{subfigure}[b]{0.52\textwidth}
        
        \includegraphics[scale=0.42]{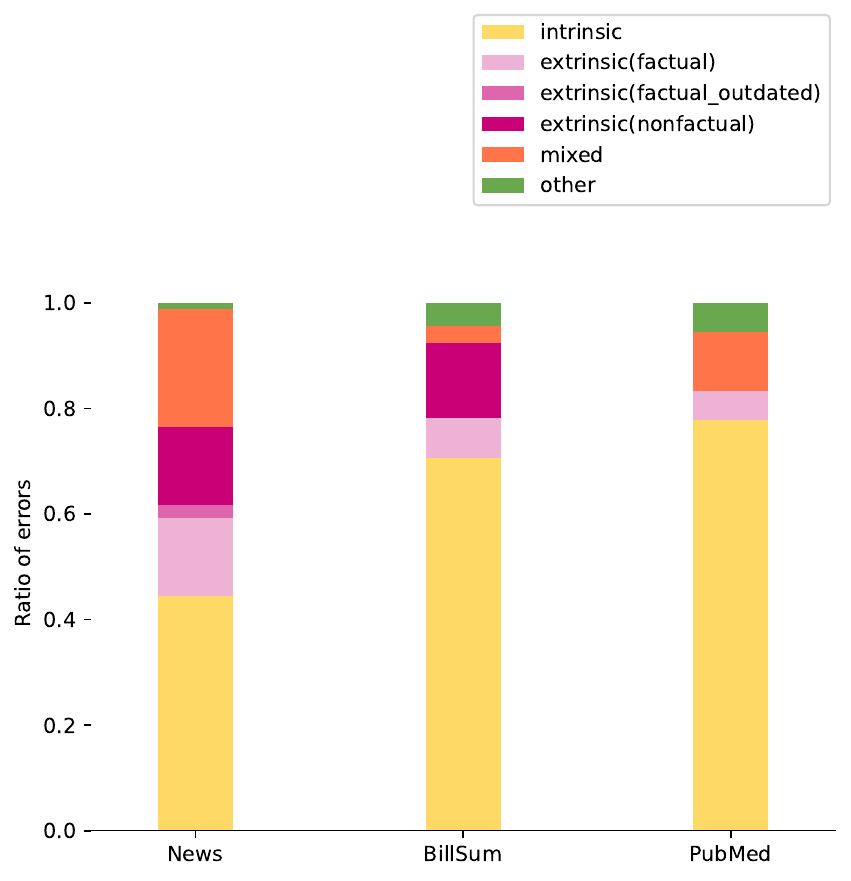}
        \caption{Distribution of error categories across domains}
        \label{fig:error-distributions}
    \end{subfigure}
    \caption{Distribution of errors and error categories across domains}
    \label{fig:overall}
\end{figure*}

To acquire manual assessments of model-generated summaries, we hire domain experts via Upwork.\footnote{Upwork is a contracting platform suited to such work because it allows hiring individuals with specific background;  \url{http://upwork.com}.}
We recruit two experts for each domain: linguistics experts for news, attorneys in civil litigation and public policy for the legal domain, and medical doctors (MDs) for the medical domain.

Our evaluation consists of two rounds. In the first round, annotators primarily assess the factual consistency of summaries in relation to the source article. We collect sentence-level annotations, instructing annotators to identify sentences with inconsistencies. The average proportion of such sentences in each domain is a key reported result. 
The inter-annotator agreement at the summary level was determined by calculating the fraction of instances where both annotators identified a summary as inconsistent with respect to the source. The agreement values are 0.80, 0.72, and 0.85 for news, billsum, and PubMed, respectively. 
We provide more details about annotation, including agreement statistics, in the Appendix \ref{sec:appendix_annotation_details}

In the second round of annotations, we categorize errors based on typology previously introduced \citep{tang2022understanding}. 
These errors include: (a) \emph{Intrinsic} errors, which  misrepresent source content, and (b) \emph{Extrinsic} errors, or ``hallucinations'', which introduce terms or concepts not in the source. Past research \citep{cao2021hallucinated} has shown that hallucinations can align with real-world knowledge and even be beneficial. 

To distinguish extrinsic errors further, we sub-categorize them into: \emph{Extrinsic nonfactual} errors, which are hallucinations inconsistent with world knowledge; and \emph{Extrinsic factual} errors, where hallucinations align with world knowledge. Additionally, considering that LLMs are trained on data up to specific points in time, we introduce
\emph{Extrinsic factual outdated} errors, which capture hallucinations that are outdated but were once in alignment with world knowledge (e.g., former presidents of countries). 
To assess the factual nature of hallucinations, annotators use online resources like Google Search and Wikipedia, in keeping with 
prior work \citep{cao2021hallucinated}.


\section{Results}

\paragraph{How prevalent are errors across domains?}
Figure \ref{fig:figure1} 
shows the average proportion of sentences marked as inconsistent (with respect to the corresponding input) in summaries generated by GPT-3.5 \cite{brown2020language} and Flan-T5 XL \cite{chung2022scaling} for three domains: News, medical, and legal. 
Perhaps surprisingly, we observe a higher prevalence of
inconsistencies for news articles, as compared to the specialized domains of medicine and law. 
While Flan-T5 introduces more errors than GPT-3.5 overall, the trends are analogous. 


\paragraph{Error categories across domains}
We next characterize the distribution of error categories in factually inconsistent summaries generated by models across the domains considererd.
Figure \ref{fig:error-distributions} reports the distribution of error categories for both models.\footnote{Model-specific distributions are in Appendix \ref{subsec:appendix_error_categories_per_model}} 
There are more extrinsic errors introduced in the news domain compared to the niche domain datasets. 
We include ``mixed'' errors for cases where errors were classified as different types (intrinsic/extrinsic) by annotators. 
The news domain has a higher frequency of such cases. 
Reviewing these, we find that they include cases where the summary both misinterprets source information and where it introduces new information. We provide examples in Appendix \ref{subsec:appendix_mixed_error}. 

An ``other'' option is available to annotators, along with a comment box for capturing miscellaneous errors. Annotator comments highlight instances where there is no clear misunderstanding but instead a misleading overall impression, such as the over-generalization of specific information in the summary


\paragraph{How extractive are summaries, and how does this relate to factuality?}
We investigate 
the relationship between extractiveness (i.e., degree of copying) and factual accuracy across domains.
Specifically, we take the proportion of 3-gram sequences in the summary that are also present in the source for each source-summary pair as a proxy measure for extractiveness. 

\begin{figure}
    \centering
    \includegraphics[scale=0.45]{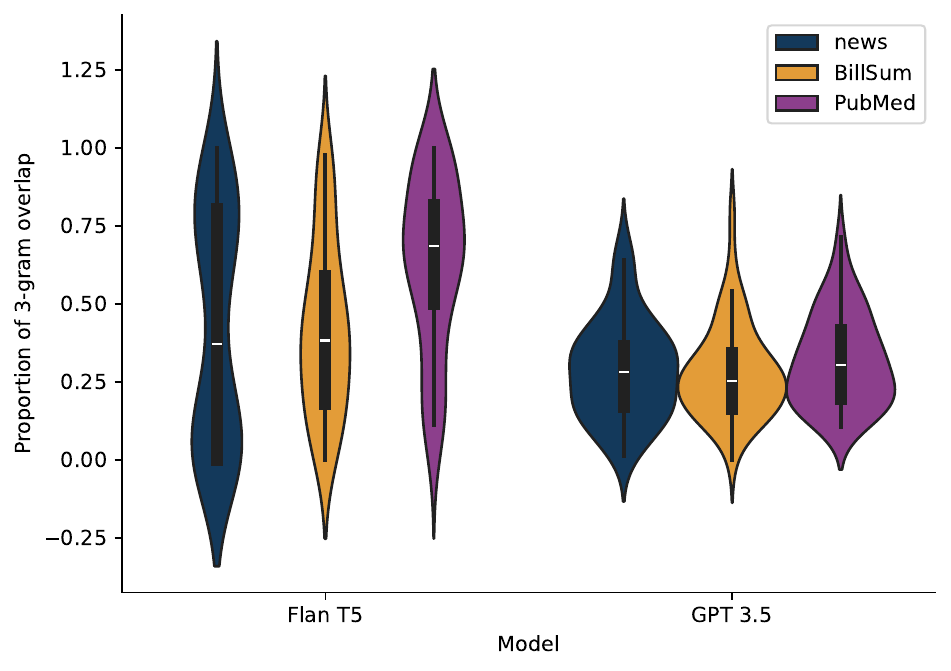}
    \caption{Proportion of 3-gram overlaps between model generated summaries and articles. We observe the most copying in the case of PubMed (especially under Flan-T5). This likely explains the greater factuality observed in this domain, and may reflect unfamiliarity with the domain (see Figure \ref{fig:fig-rougel}).}
    \label{fig:fig-copying}
   
\end{figure}

Figure \ref{fig:fig-copying} reveals that there is a 
comparable level of copying across different models and domains. 
However, 
models tend to copy more often when summarizing articles in the PubMed dataset; this could explain the lower frequency of errors in this domain, since extractive summaries are unlikely to ``hallucinate'' by definition. 
We calculated Spearman rank correlations between 3-gram overlaps and factuality scores for article-summary pairs.
The correlations for the news, billsum, and PubMed domains are 0.61, 0.38, and 0.16 respectively. 

\begin{table*}[tbh!]
\small
    \centering 
    \begin{tabular}{l c c c c c c}
         \toprule
         
        \textbf{Domain} &  \textbf{QAFactEval} & \textbf{QuestEval} & \textbf{SummC-ZS} & \textbf{SummaC-Conv} \\
        \midrule
        News & 0.58 & 0.45 & 0.47 & 0.59 \\
        BillSum & 0.27 & 0.15 & 0.23 & 0.30 \\
        Pubmed & 0.09 & -0.03 & 0.11 & 0.06 \\
    
    \bottomrule
    \end{tabular}
    \caption{Performance of automated factuality metrics across domains. We report the spearmanrank correlation between the average proportion of inconsistent sentences and the predicted scores by the automated metrics.}
    \label{tab:automated_metrics}
\end{table*}

\paragraph{Domain representation in pretraining corpora and its relation to factuality.}

\begin{figure}
    \centering
    \includegraphics[scale=0.55]{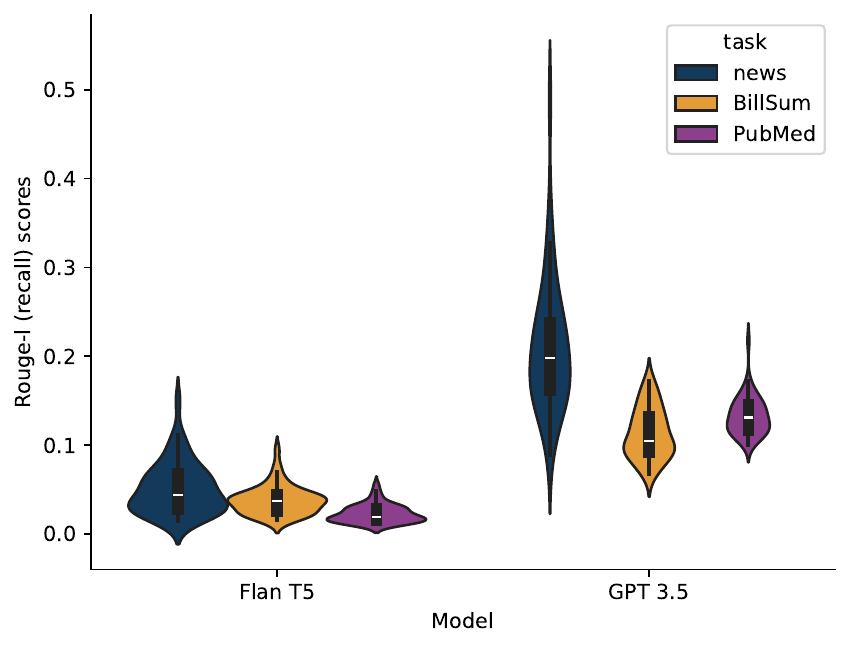}
    \caption{ROUGE-L recall scores of original articles in comparison with LLM-generated documents to measure domain exposure during pretraining. Models show higher familiarity with news topics, which may lead to the inclusion of unsupported content in summaries.}
    \label{fig:fig-rougel}
   
\end{figure}

One possible explanation for the higher proportion of factual errors in news datasets compared to specialized domains is that general news has greater representation in the training data.
As a proxy to measure model exposure to articles belonging to these domains we prompt LLMs to generate overviews of articles based on titles only (headlines for news articles, bill titles for billsum, and study titles for PubMed). 


We use the template ``Generate a comprehensive overview of the following topic: [title]'' to generate text for each article title, assessing LLMs' memorization. We speculate that increased exposure to an article topic in training data should enable LLMs to reproduce more content present in the original article (as seen with popular celebrities/events, for instance). We assess information overlap between the generated text and original article using ROUGE-L recall, favoring it over embedding based metrics because it emphasizes longest common subsequences based on exact word matches, which makes it suitable for measuring memorization. 
This is also preferable for content containing specialized terminology like PubMed abstracts and legal articles.

Figure \ref{fig:fig-rougel} shows that GPT-3.5 and Flan-T5-XL have higher ROUGE-L recall scores for news, suggesting that these models have had more exposure to news topics;  
this could explain the increased extrinsic error rate in news summaries. 
Furthermore, in Appendix \ref{subsec:appendix_alt_domain}, we show similar trends using an alternative approach to measure domain representation by directly querying the pretraining corpus with article titles, and using the number of retrieved articles as a proxy for representation.


\textbf{Are existing automatic systems for factual evaluation reliable across different domains? }
Prior research has focused on creating automated metrics for evaluating factuality of generated summaries using question answering \citep{scialom2021questeval, fabbri2021qafacteval}, natural language inference (NLI; \citealt{laban2022summac}), dependency entailment\citep{goyal2020evaluating}, and classification methods \citep{kryscinski2019evaluating}. 
The performance of these metrics has been assessed almost exclusively on evaluation benchmarks comprising model-generated summaries annotated for factuality in the news domain \citep{kryscinski2019evaluating, wang2020asking, huang2020have, maynez2020faithfulness, pagnoni2021understanding, cao2021cliff, goyal-durrett-2021-annotating, cao-etal-2022-hallucinated}. 
The effectiveness of such automated factuality metrics outside of news is underexplored. 

To address this, we use our annotated dataset to examine the performance of QAFactEval \citep{fabbri2021qafacteval}, QuestEval \citep{scialom2021questeval} and SummaC variations \citep{laban2022summac} across all three domains. 
The results in Table \ref{tab:automated_metrics} reveal that automated metrics struggle when applied to niche domains. 
We note that the lower scores observed for 
PubMed could be due to the scarcity of observed errors in this dataset, 
which makes it challenging to 
reliably evaluate its performance.

\section{Conclusions}
\label{section:conclusions}

We analyzed zero-shot summarization abilities of two LLMs, focusing on factuality. Surprisingly, inaccuracies were \emph{more likely} to be introduced in summaries of news articles compared to legal and biomedical domains.
Specifically, in this domain we observed 
more extrinsic errors---i.e., hallucinations of content not mentioned in the source---whereas errors in specialized domains 
were typically related to an apparent ``misunderstanding'' of concepts in the source. 

We hypothesize that the discrepancy could result from a higher proportion of news articles in the model's pretraining data, supported by preliminary evidence. Additionally, we observed lower Spearman rank correlations between automated metrics and human annotations in specialized domains compared to news articles, highlighting the necessity for manual evaluations or the development of new metrics for diverse benchmarks.


\section*{Limitations}
This work has a few important limitations. 
The main challenge in achieving a comprehensive evaluation is the cost involved in hiring domain experts. For news domain, we hire proofreaders and linguists at an average hourly rate of \$30 USD/hr. For billsum, we hire attorneys at \$40 USD/hr, and for pubmed, we hire doctors at \$50 USD/hr. The total cost of annotating 100 article-summary pairs across the three domains amounts to approximately \$3000 USD, making scalability of the annotations challenging.

We evaluated only two (representative) LLMs; it is possible that other models would show different patterns in behaviour. 
Another limitation of this work is that we used only a single prompt to generate summaries; although similar to a previously evaluated prompt \citep{goyal2022news} it is unclear how choice of prompt might interact with factuality of outputs across domains.

\bibliography{acl}

\appendix

\section{Appendix}
\label{sec:appendix}

\begin{table*}[tbh!]
    \centering
    
    \begin{tabularx}{\textwidth}{l c c c}
        \toprule
          & \textbf{News} &\textbf{Billsum} & \textbf{{pubmed}}   \\
         \midrule
         Avg number of source article sentences  &26.44 & 78.41 & 79.95 \\
         Avg number of summary sentences & 3.43 & 3.59 & 4.01 \\
         Avg number of inconsistent summary sentences & 0.44 & 0.38 & 0.16 \\
         \bottomrule
         
    \end{tabularx}
    \caption{Data statistics of average number of sentences in the source, summary found in the sampled data. We also include the average number of inconsistent sentences found in summaries of respective domains}
    \label{tab:appendix_data_stats}
\end{table*}
\subsection{Data Statistics}
This section presents additional data statistics in Table \ref{tab:appendix_data_stats}, including the average number of sentences in both summaries and source articles across various domains, offering context for comparisons.
\label{sec:appendix_summary_stats}
\begin{table}[h!]
    \centering
    
    \begin{tabularx}{0.5\textwidth}{l c c c}
        \toprule
         \textbf{Domain} & \textbf{Sentence} &\textbf{Category} & \textbf{Summary}   \\
         \midrule
         News & 0.91 (0.65) & 0.86 (0.45) & 0.8 (0.56) \\
         Billsum & 0.79 (0.17) & 0.78 (0.17) & 0.72 (0.37) \\
         Pubmed & 0.93 (0.11) & 0.92 (0.1) & 0.85 (0.15) \\
         \bottomrule
         
    \end{tabularx}
    \caption{We present inter-annotator agreement metrics for sentences, categories and summaries across diverse domains. Cohen's kappa scores are enclosed in parentheses for each level of annotation, often reflecting lower values. This is primarily attributed to substantial skew in error labels within the dataset, resulting in increased expected chance agreement and consequently lower kappa scores. }
    \label{tab:appendix_ann_aggr}
\end{table}

\subsection{Annotation Details}
\label{sec:appendix_annotation_details}
We recruited annotators on the Upwork platform and selected two domain experts for each task. 
In the first round, annotators identified sentences in the summary that were inconsistent with the source. 
The agreement at the summary level includes all cases where both annotators marked at least one sentence in the summary as inconsistent. 
At the sentence level, we calculated agreement as a function of the fraction of instances in which annotators marked the same sentence within a summary as being inconsistent with the source.
We calculate agreement for the error categories by considering the pre-defined error types chosen by each annotator. Notably the datasets, particularly pubmed, has an imbalance due to the dataset's significant skew in error labels, resulting in a higher expected chance agreement and lower Cohen's kappa scores. Therefore, we provide the average inter-annotator agreement and Cohen's kappa scores in the table \ref{tab:appendix_ann_aggr}

\begin{figure*}
    \centering
    \begin{subfigure}[b]{\textwidth}
       
        \includegraphics[scale=0.53]{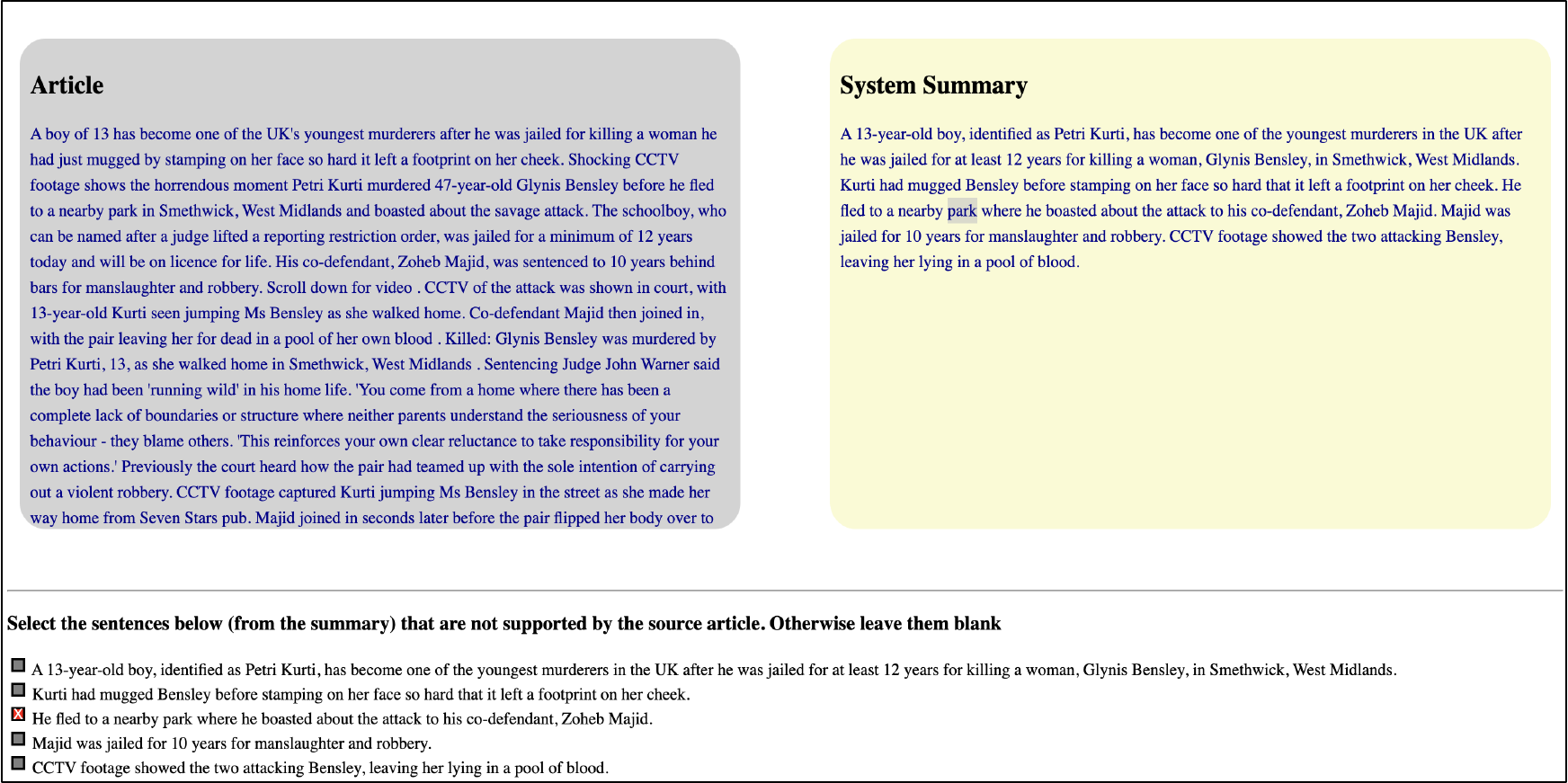}
        \caption{First round example annotation where the third sentence was marked as inconsistent . }
        \label{fig:appendix_ann_error_round1}
    \end{subfigure}

    \bigskip
    \begin{subfigure}[b]{\textwidth}
       
        \includegraphics[scale=0.51]{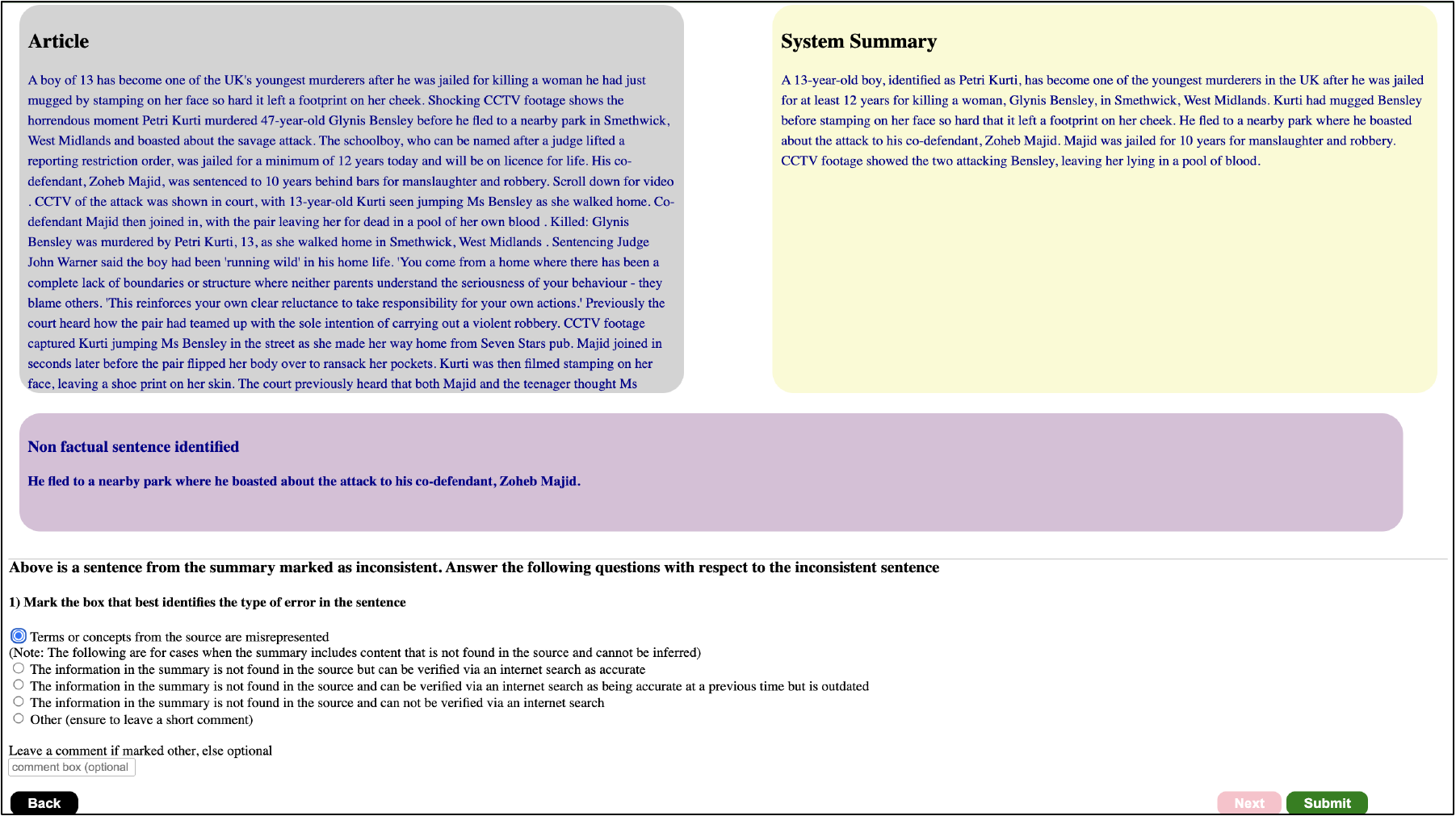}
        \caption{Second round of annotation where the annotator marked the category for the inconsistent sentence}
        \label{fig:appendix_ann_error_round2}
    \end{subfigure}
    \caption{Annotation interface with questions asked and example annotation on both round of annotations}
    \label{fig:ann_error_round1}
    
\end{figure*}
\subsection{Inconsistent summary annotation}
In the first annotation round we asked annotators to mark sentences with unsupported information, i.e., 
any information not explicitly found in the source, and which could not readily be inferred from the source alone. An example is shown in figure \ref{fig:appendix_ann_error_round1}

\subsection{Error category annotation}

In the second round of annotation, we asked annotators to categorize errors identified in the first round. 
The options provided are shown in Figure \ref{fig:appendix_ann_error_round2}.
We map the options to categories as follows 

(a) \textit{terms or concepts from the source are misrepresented} are mapped to intrinisc errors  

(b) \textit{The information in the summary is not found in the source but can be verified via an internet search as accurate} is mapped to extrinsic (factual) errors 

(c) \textit{The information in the summary is not found in the source and can be verified via an internet search as being accurate at a previous time but is outdated} is mapped to extrinsic(factual, outdated) and 

(d) \textit{The information in the summary is not found in the source and can not be verified via an internet search} is mapped to extrinsic(nonfactual)

3) \textit{Other} with a mandatory comment.

An example of this round is displayed in Figure \ref{fig:appendix_ann_error_round2}

\subsection{Mixed errors}
\label{subsec:appendix_mixed_error}
\begin{figure*}[t!]
    \includegraphics[scale=0.45]{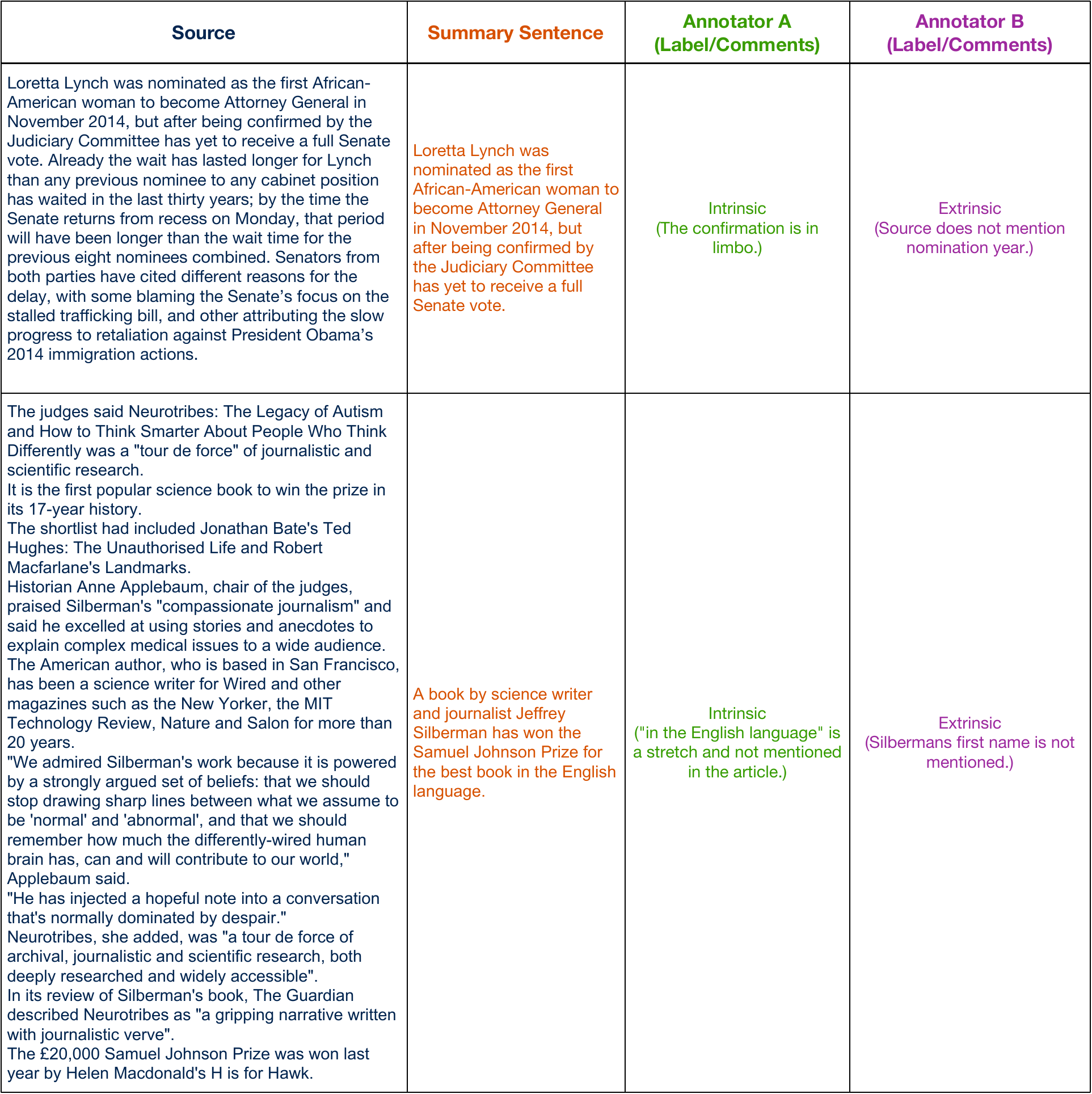}
    \caption{ Examples of sentences annotated with different categories in the news dataset by annotators along with
comments provided.}
    \label{fig:appendix_mixed_news}
\end{figure*}

We highlight some examples of the mixed error category annotations in Figure \ref{fig:appendix_mixed_news}

\subsection{Error categories per model}
\label{subsec:appendix_error_categories_per_model}
\begin{figure*}
    \begin{subfigure}[b]{0.52\textwidth}
       
        \includegraphics[scale=0.5]{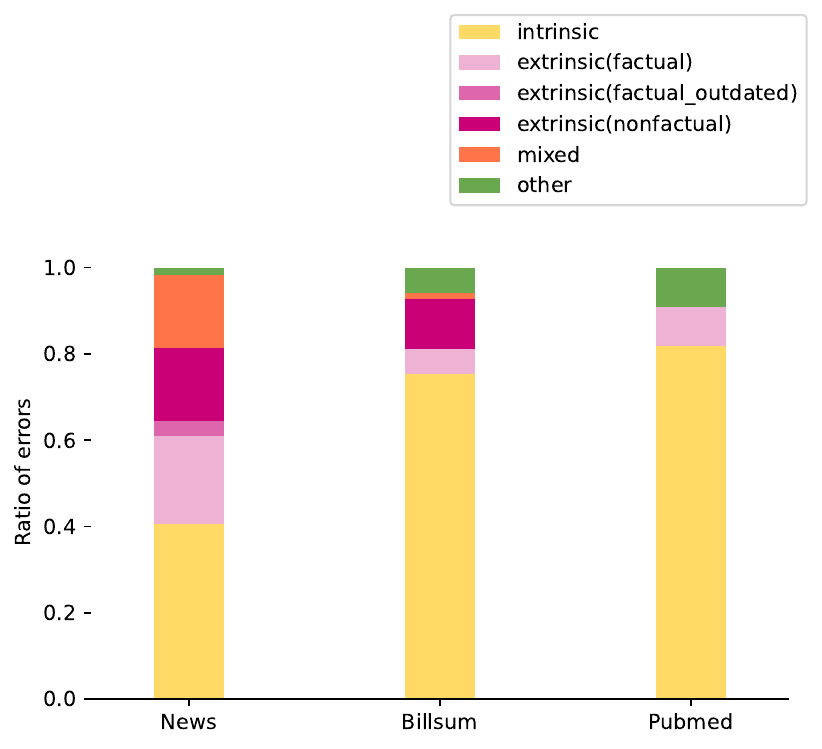}
        \caption{Distribution of error categories across domains \newline for GPT-3.5 model summaries}
        \label{fig:errorc_gpt}
    \end{subfigure}
    \hfill
    \begin{subfigure}[b]{0.52\textwidth}
        
        \includegraphics[scale=0.5]{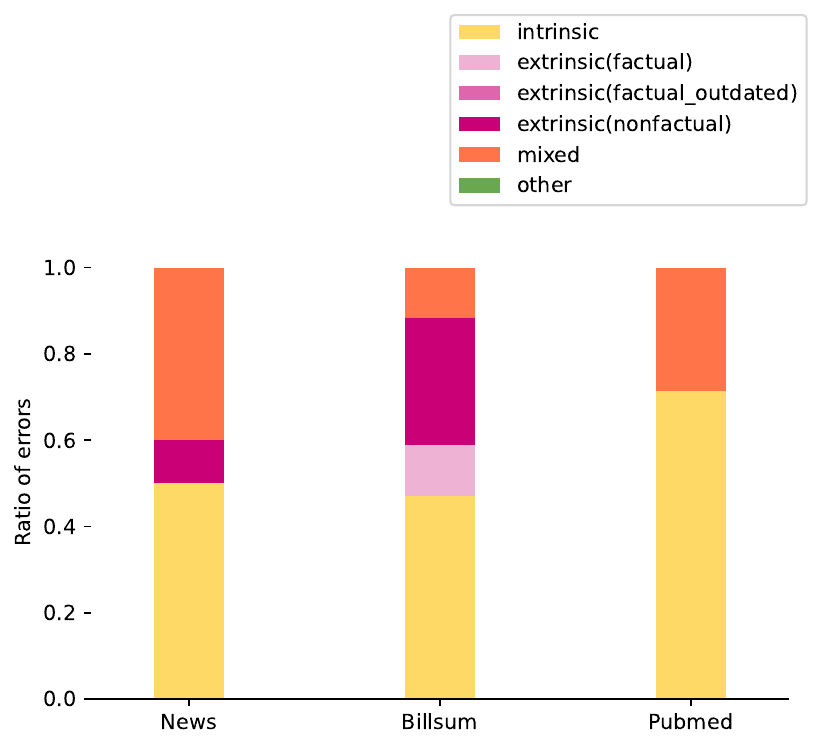}
        \caption{Distribution of error categories across domains \newline for Flan-T5-XL model summaries}
        \label{fig:errorc_flan}
    \end{subfigure}
    \caption{Distribution of error categories across domains per-model}
    \label{fig:errorc_permodel}
\end{figure*}

In Figure \ref{fig:errorc_permodel}, we present error category distributions for the Flan-T5 and GPT-3.5 models separately. Specifically, for the Flan-T5 model in the news domain, errors are typically categorized as "mixed" or marked as intrinsic and extrinsic errors, with no instances labeled as "other." For both models, the trend shows that intrinsic errors in specialized domains are equal to or higher than those in the news domain.

\begin{figure}
    \includegraphics[scale=0.55]{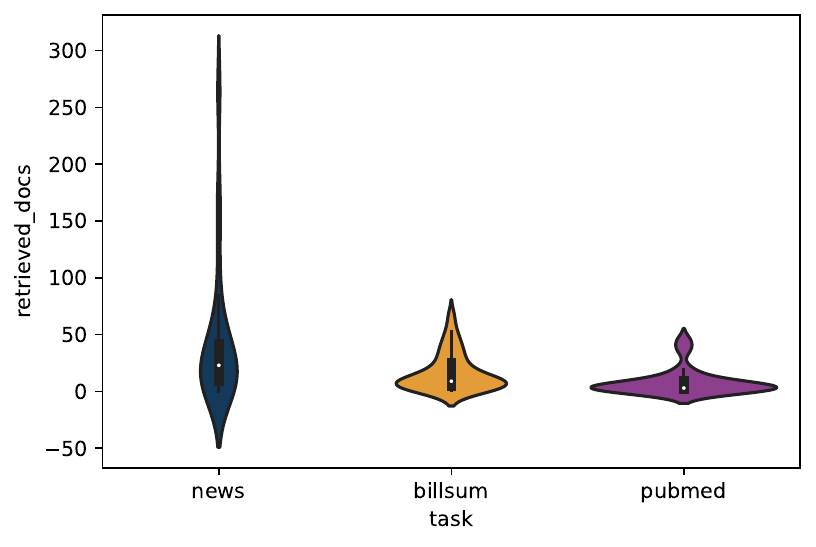}
    \caption{C-4 dataset search results for queries on news, billsum and pubmed articles. The retrieval results show that there is more representation of news articles in the C4 dataset.}
    \label{fig:c4}
\end{figure}
\subsection{Alternative method for domain representation}
\label{subsec:appendix_alt_domain}
As an alternative method for evaluating domain representation and its relation to factuality, we use the C4 dataset to query article titles. 
C4 is a large dataset derived from the the Common Crawl web corpus.\footnote{\url{https://commoncrawl.org}} 
It was used to train the T5  Transformer models \citep{raffel2020exploring}. 
The number of relevant articles found for each title serves as a proxy for article representation in the training data. 
We use a C4 search tool to query the C4 dataset.\footnote{\url{https://c4-search.apps.allenai.org/}} 
Queries for each article are manually designed using key terms from the article title with the ``AND'' condition.

Figure \ref{fig:c4} demonstrates that queries for news domain retrieved more articles in the C4 dataset compared to Billsum and Pubmed articles. 

\subsection{Model Details}
We use the default decoding parameters to generate text from GPT-3.5 and Flan-T5-XL. We use the Huggingface Transformers library \footnote{\url{https://huggingface.co/}} to implement Flan-T5-XL.

\end{document}